# Deep Learning for End-to-End Automatic Target Recognition from Synthetic Aperture Radar Imagery

Hidetoshi FURUKAWA†

† Toshiba Infrastructure Systems & Solutions Corporation
1 Komukaitoshiba-cho, Saiwai-ku, Kawasaki-shi, Kanagawa, 212–8581 Japan
E-mail: †hidetoshi.furukawa@toshiba.co.jp

**Abstract** The standard architecture of synthetic aperture radar (SAR) automatic target recognition (ATR) consists of three stages: detection, discrimination, and classification. In recent years, convolutional neural networks (CNNs) for SAR ATR have been proposed, but most of them classify target classes from a target chip extracted from SAR imagery, as a classification for the third stage of SAR ATR. In this report, we propose a novel CNN for end-to-end ATR from SAR imagery. The CNN named verification support network (VersNet) performs all three stages of SAR ATR end-to-end. VersNet inputs a SAR image of arbitrary sizes with multiple classes and multiple targets, and outputs a SAR ATR image representing the position, class, and pose of each detected target. This report describes the evaluation results of VersNet which trained to output scores of all 12 classes: 10 target classes, a target front class, and a background class, for each pixel using the moving and stationary target acquisition and recognition (MSTAR) public dataset.

## 1. Introduction

Synthetic aperture radar (SAR) transmits microwaves and generates imagery using microwaves reflected from objects, under all weather, day and night conditions. However, it is difficult for a human to recognize a target from SAR imagery since there is no color information and the shape reflected from a target changes. Therefore, automatic target recognition (ATR) from SAR imagery (or image) has been studied for many years.

The standard architecture of SAR ATR consists of three stages: detection, discrimination, and classification. Detection: the first stage of SAR ATR detects a region of interest (ROI) from a SAR image. Discrimination: the second stage of SAR ATR discriminates whether an ROI is a target or non-target region, and outputs the discriminated ROI as a target chip. Classification: the third stage of SAR ATR classifies target classes from a target chip.

In recent years, methods using convolution neural network (CNN) [1]–[4] have been successful in the classification of image recognition. Similarly, CNNs for SAR ATR have been proposed. On the moving and stationary target acquisition and recognition (MSTAR) public dataset [5], the target classification accuracy of the CNNs [6]–[9] exceeds conventional methods (support vector machine, etc.). However, most of CNNs for SAR ATR classify target classes from a target chip extracted from SAR image but do not classify multiple targets or a target chip (or SAR image) of an arbitrary size. In addition, a CNN for target classification can output score or probability of each class as classification result, but it is difficult for a human to verify the classification result.

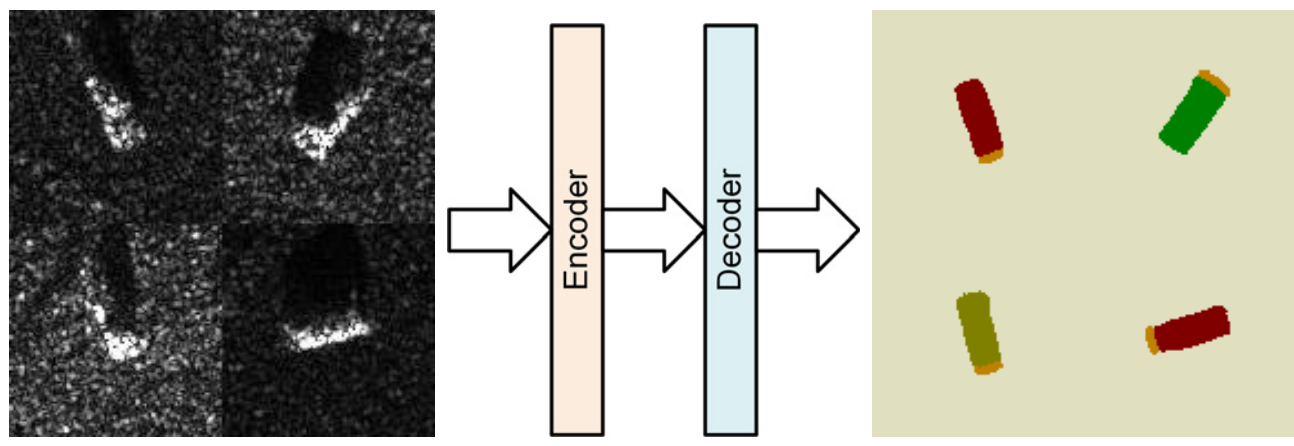

(a) Input. (b) VersNet. (c) Output.

Fig. 1 Illustration of input and output of proposed CNN. The CNN named VersNet performs automatic target recognition of multi-class / multi-target in variable size SAR image. In this case, the input is a single image with three classes and four targets (upper left and lower right targets are the same class). VersNet outputs the position, class, and pose (front side) of each detected target.

We propose a new CNN which inputs a SAR image of variable sizes with multi-target and outputs a SAR ATR image.

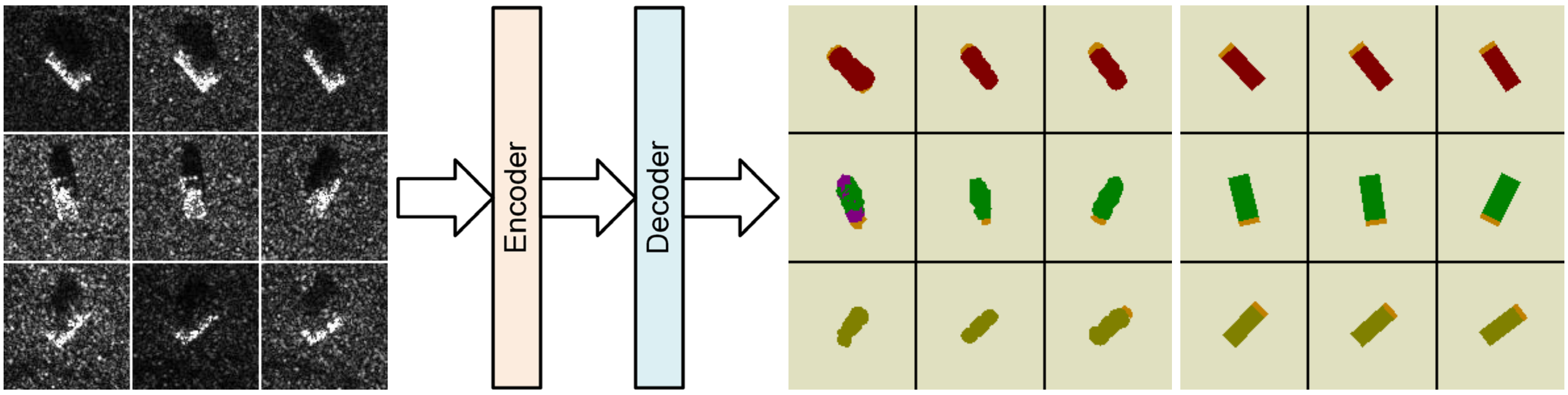

(a) Input data $X_n$. (b) VersNet. (c) Output data $Y_n$. (d) Supervised data $D_n$.

Fig. 2 Illustration of training for proposed CNN.

## 2. Related Work

In segmentation of image recognition giving classification label for each pixel of an image, methods using CNN [10]–[12] show a good performance in recent years. For SAR image, segmentation of a target region which reflected from a target, and a shadow region which not reflected from a target by radar shadow is performed. The CNN [13], WD-CFAR [14], and other methods [15]–[17] have been proposed for segmentation of a SAR image. The reference [18] describes manually generating the segmentation of target and shadow regions as ground truth. Generally, for segmentation using CNN, supervised learning is performed using label images corresponding to input images, but the difficulty of the generation of label images for SAR ATR is a problem for applying this method. In response to this problem, the reference [13] describes the CNN which trained to output a contour using the contour data of target and shadow regions generated by computer graphics as ground truth.

In contrast, our proposed CNN performs target detection, target classification, and pose estimation by segmentation.

## 3. Proposed Method

A proposed CNN named verification support network (VersNet) inputs an arbitrary size SAR image with multiple classes and multiple targets, and outputs the position, class, and pose of each detected target as a SAR ATR image.

Figure 1 shows the outline of VersNet for end-to-end SAR ATR. VersNet is a CNN composed of an encoder and a decoder. The encoder of VersNet extracts features from an input SAR image. The decoder converts the features based on the conversion rule in the training data and outputs it as a SAR ATR image.

Here, we define the end-to-end SAR ATR as a task of supervised learning. Let $\{(X_n, D_n), n = 1, ..., N\}$ be the training dataset, where $X_n = \{x_i^{(n)}, i = 1, ..., |X_n|\}$ is SAR image as input data, $D_n = \{d_i^{(n)}, i = 1, ..., |D_n|, d_i^{(n)} \in \{1, ..., N_c\}\}$ is label image for $X_n$, which is the supervised data of VersNet output data $Y_n = f(X_n; \theta)$. The values of $|X_n|$ and $|D_n|$ represent the number of pixels (vertical × horizontal) of SAR and label image, respectively. When $d_i^{(n)}$ is 1, it represents a background class, and when $d_i^{(n)}$ is 2 or more, it indicates a corresponding target class. Let $L(\theta)$ be a loss function, the network parameters $\theta$ are adjusted using training data so that the output of loss function becomes small.

Table 1 Dataset. The training and testing data contain respectively 2747 and 2420 target chips from the MSTAR.

| Class | Training data | Testing data |
|---|---|---|
| 2S1 | 299 | 274 |
| BMP2 (9563) | 233 | 195 |
| BRDM2 | 298 | 274 |
| BTR60 | 256 | 190 (1) |
| BTR70 | 233 | 196 |
| D7 | 299 | 274 |
| T62 | 299 | 273 |
| T72 (132) | 232 | 196 |
| ZIL131 | 299 | 274 |
| ZSU234 | 299 | 274 |
| Total | 2747 | 2420 |

## 4. Experiments

### 4.1 Dataset

For training and testing of VersNet, we used the ten classes data shown in Table 1 from the MSTAR [5]. The dataset contains 2747 target chips with a depression angle of 17° for the training and 2420 target chips with a depression angle of 15° for the testing. As described later in Appendix 1, five target chips of target class BTR60 for testing data were excluded.

Of course, label images for segmentation do not exist in the MSTAR dataset. Therefore, we create label images for VersNet. Figure 2(d) shows samples of label images. The label images have all 12 classes: 10 target classes, a target front class, and a background class.

(1) : Five target chips were excluded. Appendix 1 shows the details.

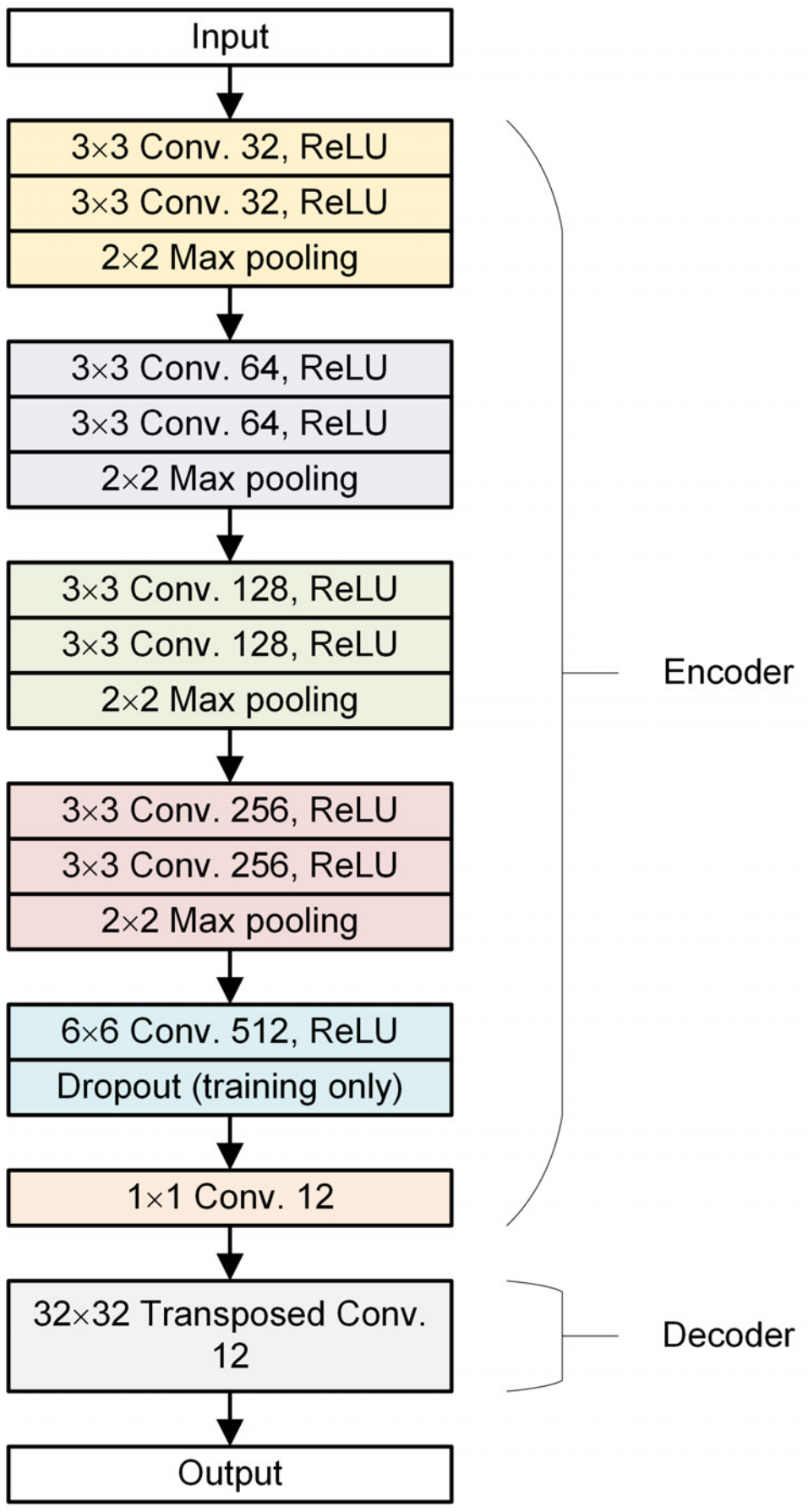

Fig. 3 Detail architecture of VersNet for experiments. The VersNet refers to the fully convolutional network called FCN-32s.

### 4.2 VersNet: Proposed CNN

Figure 3 shows a detailed architecture of VersNet for experiments. The encoder of the VersNet consists of four convolution blocks and two convolution layers. The convolution block contains two convolution layers of kernel size 3×3 and a pooling layer similarly to VGG [19]. The activation function of all convolutions except the final convolution uses rectified linear unit (ReLU) [20]. Dropout [21] is applied after a convolution of kernel size $6 \times 6$. Batch normalization [22] is not applied. The decoder of the VersNet consists of a transposed convolution [23] that performs 16 times upsampling.

As the loss function, we use cross entropy expressed by

$$L(\theta) = -\sum_{x} p(x) \log q(x). \tag{1}$$

For the optimization of the loss function, we use stochastic gradient descent (SGD) with momentum.

Since the VersNet is a CNN without fully connected layers called fully convolutional network (FCN) [10], even if training is done with small size images, the VersNet can process SAR images of arbitrary size.

### 4.3 Classification Accuracy

First, we show results of classification accuracy.

Table 2 Classification accuracy of testing data. The overall accuracy is 99.55%.

| Class | Accuracy (%) | | |
|---|---|---|---|
| 2S1 | 100.00 | | |
| BMP2 | 100.00 | | |
| BRDM2 | 98.54 | | |
| BTR60 | 98.42 | | |
| BTR70 | 98.98 | Average | Overall |
| D7 | 100.00 | 99.52 | 99.55 |
| T62 | 99.63 | | |
| T72 | 100.00 | | |
| ZIL131 | 99.64 | | |
| ZSU234 | 100.00 | | |

Table 3 Definitions of TP, FP, FN, and TN.

| True / Predicted | Condition positive | Condition negative |
|---|---|---|
| Condition positive | True positive (TP) | False positive (FP) |
| Condition negative | False negative (FN) | True negative (TN) |

Table 2 shows classification accuracy for the target chips of testing if we simply select the majority class of maximum probability for each pixel from ten target classes as the predicted class. An average accuracy of ten target classes is 99.52%, and an overall accuracy is 99.55% (2409/2420), which is almost the same as a state-of-the-art accuracy. Also, Table A·2 of Appendix shows a confusion matrix for the target chips of testing.

### 4.4 Segmentation Performance

Next, we show results of segmentation performance.

We use precision, recall, $F_1$, and intersection over union (IoU) as metrics of segmentation performance. Each metrics is given by

$$\text{Precision} = \frac{\text{TP}}{\text{TP}+\text{FP}}, \tag{2}$$

$$\text{Recall} = \frac{\text{TP}}{\text{TP}+\text{FN}}, \tag{3}$$

$$F_1 = 2 \cdot \frac{\text{precision} \cdot \text{recall}}{\text{precision}+\text{recall}}, \tag{4}$$

$$\text{IoU} = \frac{\text{TP}}{\text{TP}+\text{FP}+\text{FN}}, \tag{5}$$

where the definitions of TP, FP, FN, and TN are shown in Table 3.

Table 4 shows precision, recall, $F_1$, and IoU for all the pixels of testing. The average IoU of all 12 classes and 10 target classes are 0.915 and 0.923, respectively. Also, Table A·3 of Appendix shows a confusion matrix for all the pixels of testing.

Figure 4 shows a histogram of IoU for each image with ten target classes. A mean and a standard deviation of the IoU are 0.930 and 0.082, respectively.

Table 4 Segmentation performance for all pixels of testing. The average IoU of ten target classes is 0.923.

| Class | Precision | Recall | $F_1$ | IoU |
|---|---|---|---|---|
| Background | 0.997 | 0.999 | 0.998 | 0.996 |
| 2S1 | 0.970 | 0.953 | 0.961 | 0.925 |
| BMP2 | 0.982 | 0.942 | 0.962 | 0.926 |
| BRDM2 | 0.978 | 0.921 | 0.949 | 0.903 |
| BTR60 | 0.984 | 0.935 | 0.959 | 0.921 |
| BTR70 | 0.980 | 0.935 | 0.957 | 0.917 |
| D7 | 0.968 | 0.954 | 0.961 | 0.924 |
| T62 | 0.974 | 0.949 | 0.961 | 0.925 |
| T72 | 0.978 | 0.962 | 0.970 | 0.941 |
| ZIL131 | 0.967 | 0.953 | 0.960 | 0.923 |
| ZSU234 | 0.955 | 0.966 | 0.960 | 0.924 |
| Front | 0.885 | 0.833 | 0.858 | 0.752 |
| Average of 12 (2) | 0.968 | 0.942 | 0.955 | 0.915 |
| Average of 10 (3) | 0.974 | 0.947 | 0.960 | 0.923 |

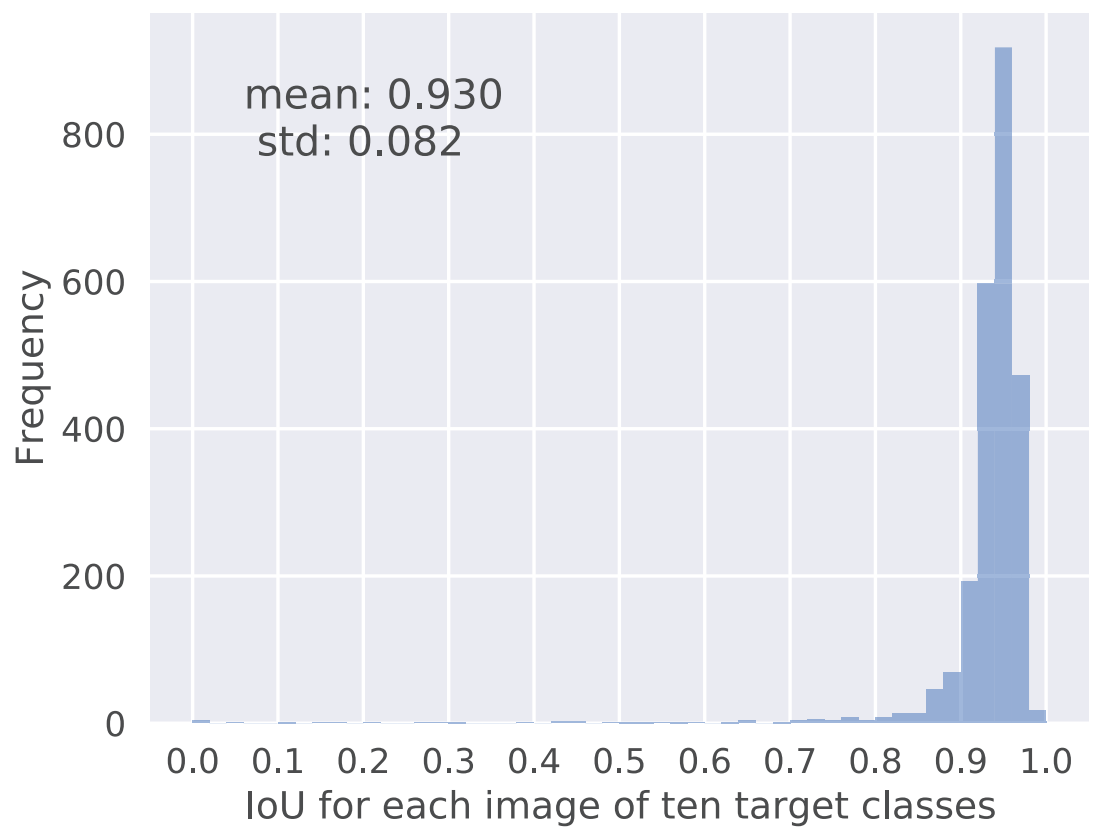

Fig. 4 Histogram of IoU.

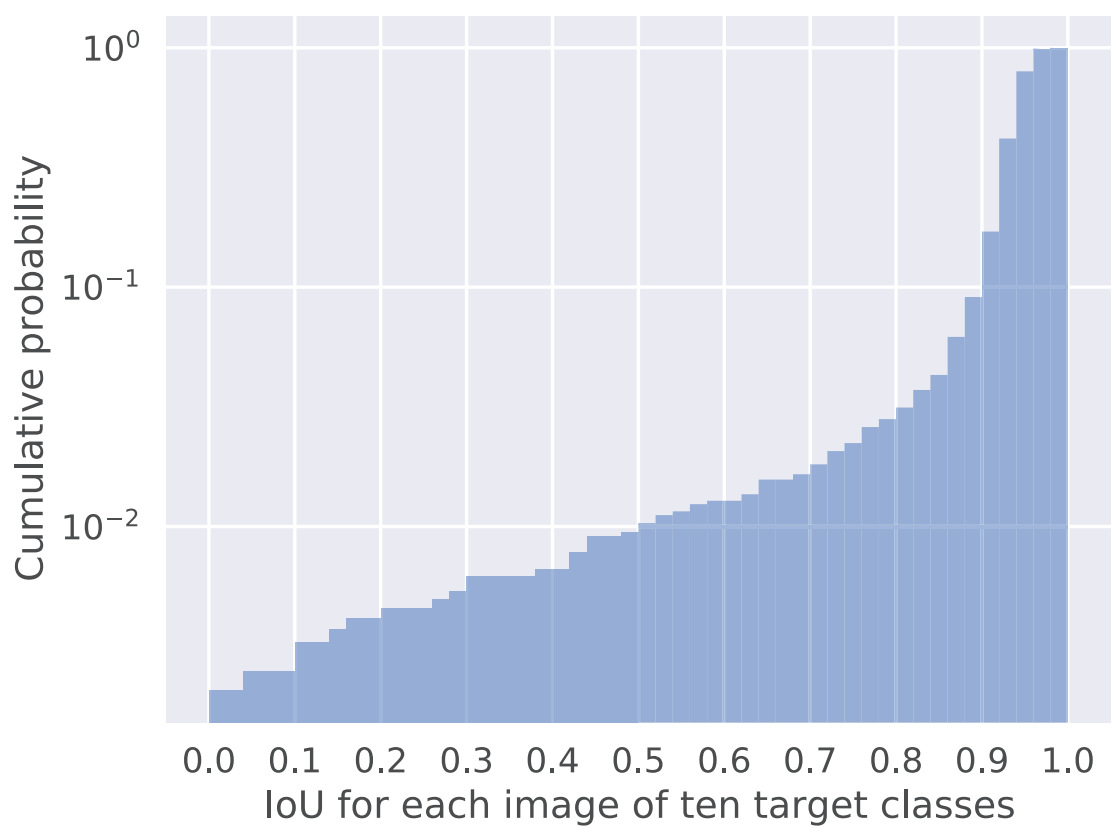

Fig. 5 Cumulative distribution of IoU.

Figure 5 shows a cumulative distribution of IoU for each image with ten target classes. The empirical cumulative distribution function P(IoU $\leq$ 0.5) and P(IoU $\leq$ 0.9) are about 0.01 and 0.1, respectively.

### 4.5 Multi-Class and Multi-Target

Finally, we show the VersNet output for multi-class and multi-target input.

Figure 6 shows input (SAR image of 10 target classes and 25 targets), output (SAR ATR image), and ground truth.

## 5. Conclusion

By applying CNN to the third stage classification in the standard architecture of SAR ATR, the performance has been improved. In order to improve the overall performance of SAR ATR, it is important not only to improve the performance of the third stage classification but also to improve the performance of the first stage detection and the second stage discrimination.

In this report, we proposed a CNN based on a new architecture of SAR ATR that consists of a single stage, i.e. end-to-end, not the standard architecture of SAR ATR. Unlike conventional CNNs for target classification, the CNN named VersNet inputs a SAR image of arbitrary sizes with multiple classes and multiple targets, and outputs a SAR ATR image representing the position, class, and pose of each detected target.

We trained the VersNet to output scores include ten target classes on MSTAR dataset and evaluated its performance. The average IoU for all the pixels of testing (2420 target chips) is over 0.9. Also, the classification accuracy is about 99.5%, if we select the majority class of maximum probability for each pixel as the predicted class.

(2) : The average of all 12 classes.
(3) : The average of ten target classes.

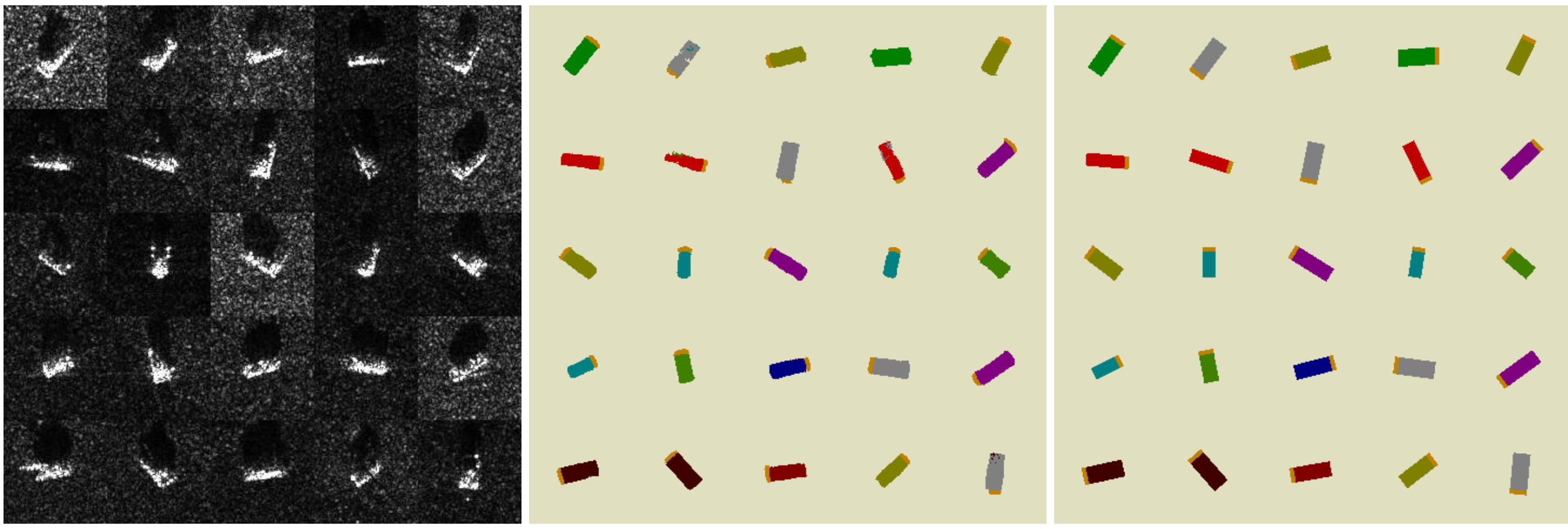

(a) Input (SAR image of 10 target classes and 25 targets).　(b) Output (SAR ATR image).　(c) Ground truth.

Fig. 6　Input (SAR image of multiple classes and multiple targets), output (SAR ATR image), and ground truth.

Table A· 1　List of target chips excluded from testing data.

| Class | Filename | Aspect angle (°) |
|---|---|---|
| BTR60 | HB04999.003 | 299.48 |
| BTR60 | HB05631.003 | 302.48 |
| BTR60 | HB04933.003 | 303.48 |
| BTR60 | HB05000.003 | 304.48 |
| BTR60 | HB03353.003 | 305.48 |

# Appendix

## 1. Excluded Target Chips from Testing Data

Table A· 1 shows a list of target chips excluded from testing data. Figure A· 1 shows inputs (target chips) and outputs (SAR ATR images) of the VersNet.

## 2. Confusion Matrix of Testing Data

Table A· 2 shows a confusion matrix for the images (target chips) of testing, and Table A· 3 shows a confusion matrix for all the pixels of testing. Each column in the confusion matrixes represents the actual target class, and each row represents the target class predicted by the VersNet.

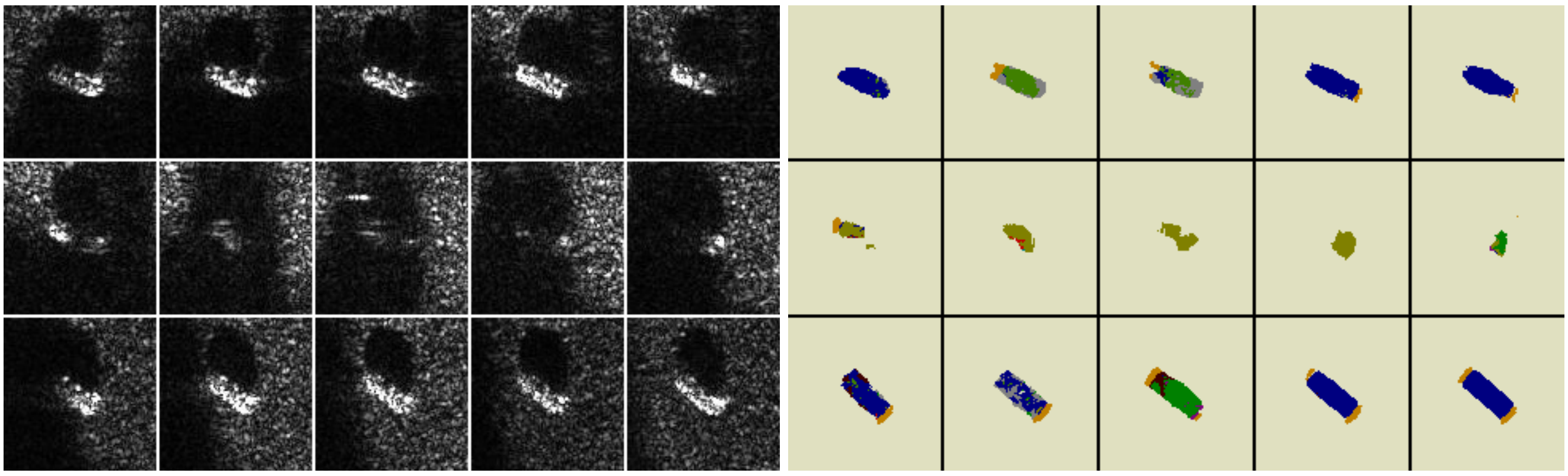

(a) Inputs (target chips of BTR60 with target aspect angle from 292° to 313°).

(b) Outputs (SAR ATR images).

Fig. A·1 Excluded target chips from testing data. As for the 2nd-row five target chips of inputs (a), the influence of the radar shadow of a certain other object appears strongly. The influence appears also in the outputs (b) of the VersNet.

Table A·2 Confusion matrix for images of testing.

| Predicted \ True | 2S1 | BMP2 | BRDM2 | BTR60 | BTR70 | D7 | T62 | T72 | ZIL131 | ZSU234 |
|---|---|---|---|---|---|---|---|---|---|---|
| 2S1 | 274 | 0 | 2 | 0 | 2 | 0 | 0 | 0 | 0 | 0 |
| BMP2 | 0 | 195 | 0 | 1 | 0 | 0 | 0 | 0 | 0 | 0 |
| BRDM2 | 0 | 0 | 270 | 0 | 0 | 0 | 0 | 0 | 0 | 0 |
| BTR60 | 0 | 0 | 0 | 187 | 0 | 0 | 0 | 0 | 0 | 0 |
| BTR70 | 0 | 0 | 0 | 0 | 194 | 0 | 0 | 0 | 0 | 0 |
| D7 | 0 | 0 | 0 | 0 | 0 | 274 | 0 | 0 | 0 | 0 |
| T62 | 0 | 0 | 0 | 1 | 0 | 0 | 272 | 0 | 0 | 0 |
| T72 | 0 | 0 | 0 | 0 | 0 | 0 | 0 | 196 | 0 | 0 |
| ZIL131 | 0 | 0 | 2 | 0 | 0 | 0 | 0 | 0 | 273 | 0 |
| ZSU234 | 0 | 0 | 0 | 1 | 0 | 0 | 1 | 0 | 1 | 274 |
| Accuracy (%) | 100.00 | 100.00 | 98.54 | 98.42 | 98.98 | 100.00 | 99.63 | 100.00 | 99.64 | 100.00 |
| | Average 99.52 | | | | | | | | | |
| | Overall 99.55 | | | | | | | | | |

Table A·3 Confusion matrix for all pixels of testing.

| Predicted \ True | Background | 2S1 | BMP2 | BRDM2 | BTR60 | BTR70 | D7 | T62 | T72 | ZIL131 | ZSU234 | Front | Precision |
|---|---|---|---|---|---|---|---|---|---|---|---|---|---|
| Background | 21032341 | 4757 | 5013 | 7127 | 3808 | 5187 | 3308 | 5611 | 3535 | 4459 | 2907 | 20339 | 99.69 |
| 2S1 | 1646 | 124249 | 169 | 1071 | 0 | 602 | 0 | 71 | 0 | 89 | 0 | 209 | 96.99 |
| BMP2 | 1080 | 5 | 99985 | 34 | 376 | 6 | 0 | 0 | 178 | 0 | 0 | 192 | 98.16 |
| BRDM2 | 1947 | 109 | 119 | 113068 | 11 | 43 | 0 | 0 | 0 | 0 | 0 | 263 | 97.84 |
| BTR60 | 1159 | 0 | 2 | 0 | 84603 | 7 | 0 | 0 | 0 | 12 | 0 | 191 | 98.41 |
| BTR70 | 1758 | 0 | 0 | 0 | 23 | 97469 | 0 | 0 | 0 | 0 | 0 | 251 | 97.96 |
| D7 | 2166 | 0 | 0 | 0 | 0 | 0 | 75195 | 15 | 0 | 9 | 3 | 301 | 96.79 |
| T62 | 1877 | 775 | 0 | 20 | 714 | 0 | 0 | 140983 | 7 | 91 | 13 | 327 | 97.36 |
| T72 | 1546 | 31 | 354 | 0 | 142 | 0 | 0 | 6 | 102533 | 0 | 0 | 202 | 97.82 |
| ZIL131 | 2701 | 12 | 2 | 803 | 2 | 0 | 1 | 123 | 0 | 112804 | 1 | 193 | 96.71 |
| ZSU234 | 2312 | 0 | 0 | 0 | 462 | 0 | 2 | 1161 | 16 | 309 | 96286 | 277 | 95.50 |
| Front | 9590 | 490 | 448 | 613 | 349 | 896 | 342 | 620 | 363 | 567 | 492 | 113784 | 88.51 |
| Recall | 99.87 | 95.26 | 94.24 | 92.12 | 93.49 | 93.53 | 95.37 | 94.88 | 96.16 | 95.32 | 96.57 | 83.34 | |
| $F_1$ | 99.78 | 96.12 | 96.16 | 94.90 | 95.89 | 95.69 | 96.07 | 96.10 | 96.98 | 96.01 | 96.03 | 85.85 | |
| IoU | 99.56 | 92.53 | 92.61 | 90.29 | 92.10 | 91.74 | 92.44 | 92.50 | 94.14 | 92.33 | 92.37 | 75.20 | |